\documentclass[10pt,twocolumn,letterpaper]{article}
\pdfoutput=1
\usepackage{cvpr}
\usepackage{times}
\usepackage{epsfig}
\usepackage{graphicx}
\usepackage{amsmath}
\usepackage{subcaption}
\usepackage[ruled,vlined, noend]{algorithm2e}
\usepackage{amssymb}
\usepackage{xcolor}
\usepackage{pdfpages}
\pagecolor{white}

\usepackage{enumitem}

\usepackage{graphicx}

\usepackage[breaklinks=true,bookmarks=false]{hyperref}

\cvprfinalcopy 

\def\cvprPaperID{****} 

\begin{document}

\title{Bonsai-Net: One-Shot Neural Architecture Search via Differentiable Pruners}

\author{Rob Geada, Dennis Prangle, Andrew Stephen McGough  \\
Newcastle University, UK\\
{\tt\small \{r.geada2, dennis.prangle, stephen.mcgough\}@newcastle.ac.uk} \\ \and 
}

\maketitle

\begin{abstract}
	One-shot Neural Architecture Search (NAS) aims to minimize the computational expense of  discovering state-of-the-art models. However, in the past year attention has been drawn to the comparable performance of na{\"i}ve random search across the same search spaces used by leading NAS algorithms. To address this, we explore the effects of drastically relaxing the NAS search space, and we present Bonsai-Net \footnote{See \texttt{https://github.com/RobGeada/bonsai-net-lite} for code and implementation details.}, an efficient one-shot NAS method to explore our relaxed search space. Bonsai-Net is built around a modified differential pruner and can consistently discover state-of-the-art architectures that are significantly better than random search with fewer parameters than other state-of-the-art methods.  Additionally, Bonsai-Net performs simultaneous model search and training, dramatically reducing the total time it takes to generate fully-trained models from scratch.
\end{abstract}
\section{Introduction}
Neural Architecture Search (NAS) is a field which contains many promising methods for developing state-of-the-art architectures. However, recent work by Yu \etal \cite{yu2019} and Li \& Talwalkar \cite{li2019} found the performance of these methods to be barely better, if not worse, than random search. Our hypothesis as to why this occurs mirrors that of Yu \etal; the search space for these methodologies is over-constrained to the extent that \textit{any} model from the search space would perform well. This is in part due to the saturation of the CIFAR-10 problem, in that all the cumulative research on best practices has leaked into the design of the search spaces. This means that the search spaces exclusively contain excellent CIFAR-10 architectures, which both eliminates any potential benefit of NAS as well as limits the generalizability of these spaces to other problem domains. 

To address these issues, we have designed a new search space that significantly relaxes the restrictions commonly seen in other NAS methods, such as to remove their implicit design biases but also explore novel design patterns not seen in other works.  Additionally, we have produced a one-shot NAS method capable of efficiently discovering state-of-the-art models within this new space via differentiable, memory-aware pruners, such as to investigate the efficacy of NAS over broader search spaces. 
 
\section{Search Space}
Our search space is based on the search space of DARTS \cite{liu2019} and PC-DARTS \cite{xu2020}, whose models are composed of stacked cells, each cell a directed graph wherein edges are tensor operations and nodes are tensor aggregations. Cells are classified as either a \textit{normal} cell, meaning that tensor dimensionality remains unchanged throughout, or as a \textit{reduction} cell, meaning that spatial dimensions are halved while the channel dimension is doubled. However, in both of these papers and in most existing NAS models, each cell $C_n$ is restricted to receiving input from the two previous cells $C_{n-1}$ and $C_{n-2}$, and each reduction or normal cell in the model is necessarily homogenous, that is, identical to every other cell of the same type. In their space, there are roughly $6\times10^{11}$ possible 4 node cells, 2 different cells per model, and 1 possible set of connections between these cells.

To relax this search space, we allow each cell to receive input from both the previous cell as well as any combination of previous cells, each node can receive input from any combination of previous nodes in the cell, and each edge can contain any combination of the operations in the operation space. Most importantly, each cell in the network is distinct, meaning each cell has a distinct connectivity and operation set. The operation space is equivalent to that of PC-DARTS: identity, 3x3 average pooling, 3x3 max pooling, 3x3 separable and dilated convolutions, and 5x5 separable and dilated convolutions. Therefore, our search space is a significantly larger superset of the DARTS space; an 8 cell model has around $3 \times 10^{29}$ possible 4 node cells, 8 different cells per model, and $254$ different sets of connections between the cells.

\section{Differentiable Pruners}
The differentiable pruner was introduced by Kim \etal \cite{kim2019} in 2019 as a means of pruning FLOPs from neural operations. The authors describe the problem of gating channels via a gate function, where the gate function has a constant derivative of 0 everywhere it is differentiable, meaning that gradient descent cannot be aware of the effects of the gate. To remedy this, the authors add a saw wave $S$ to the gate function $G$, creating a trainable gate function $P$:
\begin{align}
	G(w) &= \begin{cases}
						0 & w<0 \\
						1 & w\ge 0
					\end{cases} \\
	S(w) &= \frac{M w - \lfloor M w \rfloor}{M} \\
	P(x, w) &= (G(w) + S(w)) x
\end{align}
where $x$ is the output tensor of the operation to be pruned, and $w$ is the vector of weights that control the gating effects of the pruner. With large enough $M$, $G(w)+S(w)$ is effectively 0 or 1 everywhere while still having a constant derivative of 1 everywhere it is differentiable, as visualized in Figure 1 of the supplementary materials. This lets gradient descent track the effects of gating, and thus allows for gating to be a parameter learned simultaneous with model training. Compression is encouraged by adding a compression term to the loss function, computed as follows:
\begin{align}
	\mathcal{L}_{comp} &= \lambda \lVert c_{target} - \vec{c}_{actual} \rVert \label{eq:comp_term}
\end{align}
where $\lambda$ is the compression weight (such as to balance this term with the regression or classification loss of the model), $c_{actual}$ is the vector of the model's compression, and $c_{target}$ is the desired compression level. In the original paper, compression is computed as the ratio of the model's current FLOP count to its original FLOP count.

\subsection{Memory-Aware Pruners}
For our purposes, we adapt the differentiable pruner from a channel-wise operation to a generic binary operation, that can gate any connection within the model. The weight vectors of these adapted pruners are of size 1, meaning that the tensors passed through the pruners are either entirely preserved or entirely pruned.  In addition, we compute the memory size $s$ of each potentially pruned connection in the network, to be used in the computation of our $\vec{c}_{actual}$, which is the vector of the compression of each cell in the model:
\begin{align}
	c_{{actual}_i} &= \frac{\text{Memory Size Unpruned Operations in } C_i}{\text{Memory Size All Operations in } C_i} \\[.5em]
	&=  \frac{\sum\limits_{j=0}^{\# cnx \in C_i}{\left(s_j* (G(w_j) + S(w_j))\right)}}{\sum\limits_{j=0}^{\# cnx \in C_i}{s_j}} 
\end{align}
where $C_i$ is the $i$th cell of the model, $\# cnx$ is the number of connections in $C_i$, $s_j$ is the memory size of the $j$th connection, and $w_j$ the weight value of the pruner along that connection. Therefore, a cell that started with 8 GiB VRAM usage and then pruned down to 6 GiB would have a compression ratio of $\frac{6}{8} = 0.75$. 

The $\lambda$ term in $\mathcal{L}_{comp} $ as per equation \ref{eq:comp_term} is chosen such that the overall compression penalty does not overpower the classification or regression loss of the network; if we must encourage pruning we want to minimize its effect on the model's performance on the target task. From our experiments, best results occur when the compression loss is around 1\% of the classification loss, which typically corresponds to a  $\lambda$ term on the order of $0.01$.

The advantage of memory-aware pruners is that they allow the memory cost of a model to be a directly differentiable parameter. In our experience, GPU memory is a major limiting factor in deployment, and thus being able to directly tune memory cost during training to a variety of memory and hardware constraints is tremendously useful.

\section{NAS}
\subsection{Bonsai-Net}
Bonsai-Net is our NAS model that leverages these memory-aware differentiable pruners to discover models in our relaxed search space. Bonsai-Net creates a hypernetwork similar to DARTS, wherein the model is initialized with every possible connection within the search space, called the \textit{hyper-connected} state. This means that every cell receives the output of the $n-1th$ cell as well as the combination of every previous cell's output, each node within the cell receives the combination of each previous node, and each edge is the combination of each operation in the operation set. Pruners are put along each intra-cellular and inter-cellular connection, such that Bonsai-Net can eliminate any connection as it sees fit. The hyper-connected state is visualized in Figure 2 of the supplementary materials.

\subsection{The Bonsai Process}
Due to the immense number of connections in the Bonsai-Net hypernetwork, the entirety of the hyper-connected model cannot be fit into GPU memory at first. Instead, models are divided into groups of consecutive cells called sections, such that each section is roughly the same size and the first section can fit into GPU memory in the hyper-connected state. The model is then trained iteratively; the first section is added to the model along with a temporary classification tower (consisting of global average pooling followed by a fully-connected layer), then the model is trained and pruned until the next section can fit into memory. When a new section is added, the weights of the previous sections are preserved and the old classification tower is converted into an auxiliary tower to help bootstrap the new section.

We've likened this growing and pruning process to that of growing a bonsai tree, hence the name. Through the use of memory-aware pruners, we can directly encourage Bonsai-Net to free up GPU space for these future sections, via a compression term as per equation \ref{eq:comp_term} added to the loss function. The compression targets for this term are established at the start of training, by determining how much space each new section of the hypernetwork requires. The lengths of the pruning periods for each section are not fixed, but rather depend on the time it takes for the model to achieve the necessary compression. These intermediate training steps are included in Bonsai-Net's training allocation, and thus the total number of training epochs is fixed regardless of how long the model takes to build itself to full size. At this full size, the model trains with no compression term in the loss, meaning that any compression that does occur is solely motivated by performance. The full algorithm is detailed in algorithm \ref{alg:bonsai} below:
\begin{algorithm}
\SetAlgoLined
For all $n \in [1, \#_{sections}]$, determine the compression $c_n$ s.t. section $s_n$ can be appended to the model\;
Initialize hyper-connected section $s_0$\;
\BlankLine
\For{n in $[1, \#_{sections}]$}{
	\While {$c_{actual}>c_n$}{
		Train + prune, $c_{target}=c_n$\;
	}
	Convert classification tower to auxiliary tower\;
	Add hyper-connected section $s_n$\;
	Add new classification tower\;
}
Finish remaining train+prune epochs, $c_{target}= $ None\;
\caption{The Bonsai-Net algorithm}
\label{alg:bonsai}
\end{algorithm}
\section{Experimental Setup}
\subsection{CIFAR-10 Configuration}
The configuration for our Bonsai-Net models for CIFAR-10 is chosen to be similar to that of DARTS. Models have 8 cells, with reductions cells placed at 1/3rd and 2/3rds of model depth. The initial channel count is 36 with a batch size of 64, and the models are trained for 600 epochs with a cosine-annealed learning rate starting at 0.01. We use a drop-path value of 0.3 \cite{larsson2016}, and data is augmented with the standard set of augmentations and cutout. Compression $\lambda$ is set to $0.01$ at the start of each prune cycle, and is doubled every 16 epochs while the compression target is not met. The cellular composition as well as time spent at each section from an example run of Bonsai-Net on CIFAR-10 is given in table 3 in the supplementary materials.

\subsection{Random Search and Ablation Study}
We tested Bonsai-Net against two different levels of randomness, to both evaluate our algorithm as well as perform an ablation study. Both levels are trained for the same number of epochs with the same hyper-parameters as the Bonsai-Net models, to allow for direct comparison. The levels are as follows:

\textbf{Level 1:} Sections 1 through $n-1$ are randomly connected at the same compression level as sections 1 through $n-1$ of the Bonsai-Net model. Section $n$ is added at full size. The model is allowed to prune internal connections.  This level tests the effectiveness of iteratively building the model up to its final size via the Bonsai process versus random search.

\textbf{Level 2:}  The full model is initialized at the same compression as the fully trained Bonsai-Net model, and is not allowed to prune. This level tests the efficacy of the joint Bonsai and pruning process in generating model architectures versus pure random search. This is equivalent to the random search in the work of Yu \etal and Li \& Talkwalkar. 

\section{Results and Discussion}
\subsection{Performance and Comparisons}
\begin{table}[h]
\begin{center}
\begin{tabular}{c|cc}
& Test Acc. & Parameters\\
\hline
Bonsai-Net  & \textbf{96.65  $\pm$ 0.06\%}  & \textbf{2.95 $\pm$ 0.11 M} \\
Random-1   & 95.27  $\pm$ 0.05\% & 3.89  $\pm$ 0.12 M\\
Random-2   & 95.19  $\pm$ 0.13\% & 3.03 $\pm$ 0.01 M
\end{tabular}
\end{center}
\caption{Bonsai-Net performance, parameter counts, and total runtime versus the two levels of random search. Each configuration is tested three times, and the average of these three runs is reported. Models used 7.21 $\pm$ 0.14 GiB of GPU memory on average. Error bounds are computed as the standard error of mean.}
\label{tab:performance}
\end{table}
\begin{table}[h]
\begin{center}
\begin{tabular}{c|cc}
 \multicolumn{1}{c}{} & \multicolumn{2}{c}{Restricted Search Space} \\ 
Model & Test Acc. & $\Delta$ Random Search \\ 
\hline
NAO           		& \textbf{96.86 $\pm$ 0.17}\% 	& 0.38\% \\
ENAS			& 96.76 $\pm$ 0.10\% 		& 0.28\% \\
DARTS       		& 96.62 $\pm$ 0.23\% 		& 0.14\% \\
BayesNAS  		& 95.99 $\pm$ 0.25\%		& -0.44\% \\
Random	  		& 96.48 $\pm$ 0.18\%		& -\\
\multicolumn{3}{c}{} \\[-1em]
 \multicolumn{1}{c}{} & \multicolumn{2}{c}{Bonsai Search Space} \\
\hline
Bonsai-Net 		& 96.65 $\pm$ 0.05\% 	& \textbf{1.45\%} \\
Random	  		& 95.19 $\pm$ 0.13\%	& -\\
\end{tabular}
\end{center}
\caption{Bonsai-Net performance versus random search, compared to other NAS algorithms.  Results for other algorithms is taken from Yu \etal  \cite{yu2019}. Level-2 random search is chosen for comparison as it aligns with the random-search methodology of Yu \etal.}
\label{tab:comp_performance}
\end{table}
Table \ref{tab:performance} shows that both random searches performed significantly worse than Bonsai-Net, despite using the same amount of GPU memory and roughly similar parameter counts, which indicates the effectiveness of Bonsai-Net in discovering and training models. Bonsai-Net achieves similar mean performance to other state-of-the-art methods as per table \ref{tab:comp_performance}, while requiring 350K fewer parameters than DARTS, 650K fewer than PC-DARTS, and 1.7M fewer than ENAS. Since the search and train phases of Bonsai-Net are codependent, we report the total time from the commencement of the algorithm to the completion of training, which is typically around 3.3 GPU days on an NVidia 1080Ti, nearly twice as fast as the total runtime of second-order DARTS on the same GPU (6.5 GPU days).  

Additionally, table \ref{tab:comp_performance} demonstrates that de-constricting the search space has indeed removed some of the design bias of the restricted space; the average randomly-selected model in the Bonsai search space performs 1.29 percentage points worse than the average randomly-selected model in the constricted search space of other NAS algorithms. However, despite the de-constricted space, Bonsai-Net still discovers state-of-the-art architectures. While future work is necessary to compare how other NAS methods perform in the Bonsai search space, implementing the other methods into a search space that does not enforce cellular homogeneity may prove to be impossible due to VRAM constraints.

\subsection{Observations} 
While observing the pruning patterns of our models, consistent `choices' became noticeable. These seem to indicate that the architectural decisions being made are deliberate, learned choices rather than the result of random chance:
\begin{itemize}
\setlength\itemsep{0em}
\item Models seem to have a preferential ranking of operations, demonstrated by a consistent occurrence frequency of various operations in different types of cells. In normal cells, the most frequently chosen operation by far is identity operations, followed by separable convolutions, then max pooling. Dilated convolutions are rarely chosen, around a dozen per model, while average pooling has never been chosen in any recorded run. Operation counts are shown in Figures 3 and 4 of the supplementary materials.
\item Average pooling operations are always pruned out of the model entirely. Discovering why this is the case presents an interesting avenue for future investigation.
\item At a higher organizational level, Bonsai-Net tends to build ResNet-esque \cite{he2015} patterns within cells, favoring edges that are the summation of identity and separable convolution operations. This is one of the most common learned designs for edges within Bonsai-Net, and is a configuration that is only possible within our search space. Specific examples are shown in Figure 5 of the  supplementary materials.
\item Cells prefer to receive connections from closer cells rather than further ones. The most frequently chosen input for cell $C_n$ is $C_{n-1}$, then $C_{n-2},$ and so on. 
\end{itemize}
 
\section{Conclusion}
We have created an expanded Neural Architecture Search space, and have designed a one-shot NAS method to produce state-of-the-art results over this new space. Our search space presents a more realistic starting point for NAS; when approaching a novel problem, the intuition and design experience that is encoded into the restricted search spaces of other NAS algorithms does not exist, and thus techniques must be able to operate on significantly broader search spaces. Our results demonstrate that our search space is significantly less constricted than that of other NAS methods, but over this expanded and more realistic space our one-shot NAS algorithm Bonsai-Net still produces state-of-the-art results, while generating fully-trained models in a much faster time and with fewer parameters than other leading algorithms.
{\small
\bibliographystyle{ieee_fullname}
\bibliography{bib}
}
\newpage



\section*{Supplementary material}

# Bonsai-Net: Supplementary Materials

## 1 Pruner Operation

$G(w) + S(w)$ from equation 3 in the main paper is shown below:

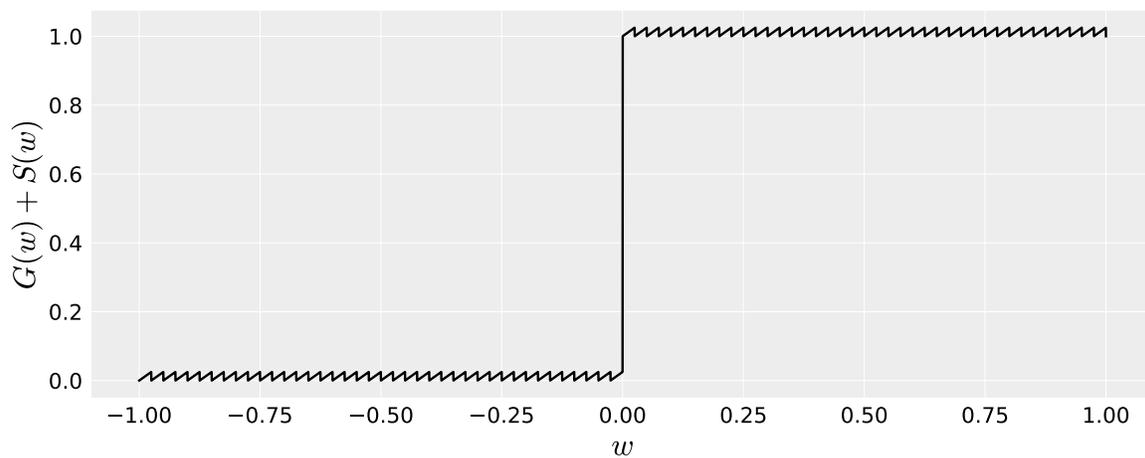

Figure 1: $G(w) + S(w)$, shown at an unrealistically small $M = 40$ for clarity. Larger $M$ values (typically around $1e5$ in actual models) reduce the magnitude of the saw component to imperceptiblity. Notice how the saw function introduces a consistent gradient to the gate function without drastically affecting the function's overall shape.

## 2 Hyper-Connectivity

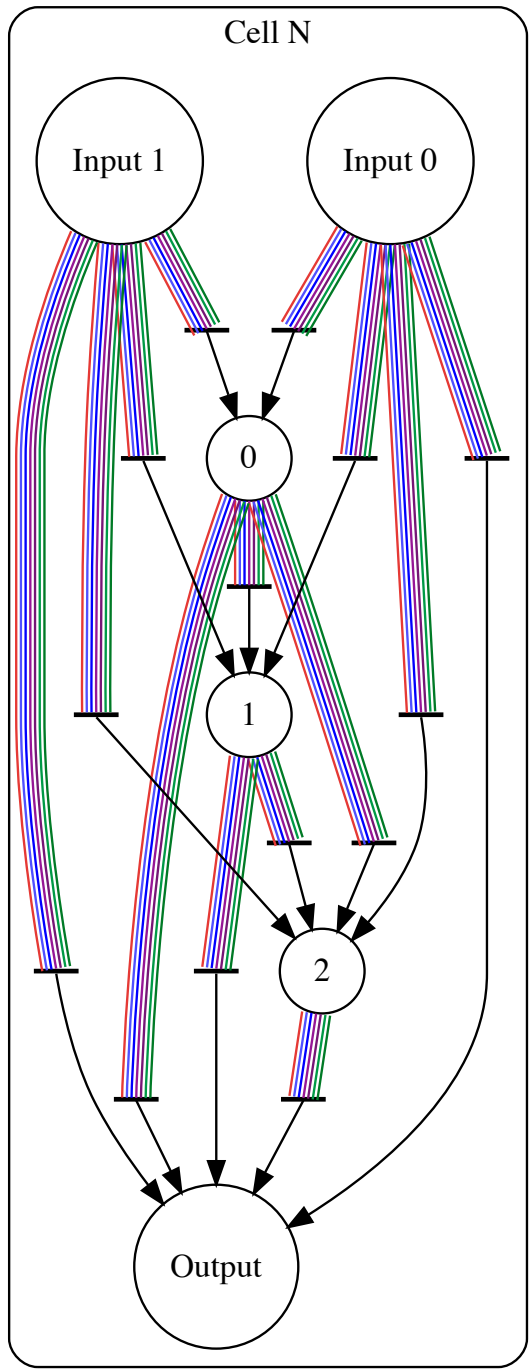

(a) A hyperconnected four node cell, the same size as the cells in the Bonsai-Net runs described in the main paper. Pruners are represented by horizontal black boxes. Input 0 is the permanent, hard-coded input from the previous cell, while Input 1 is the prunable input from each previous cell. Between each node is each of the possible operations in the search space, all of which are independently passed through pruners before being summed together and inputted into their target node.

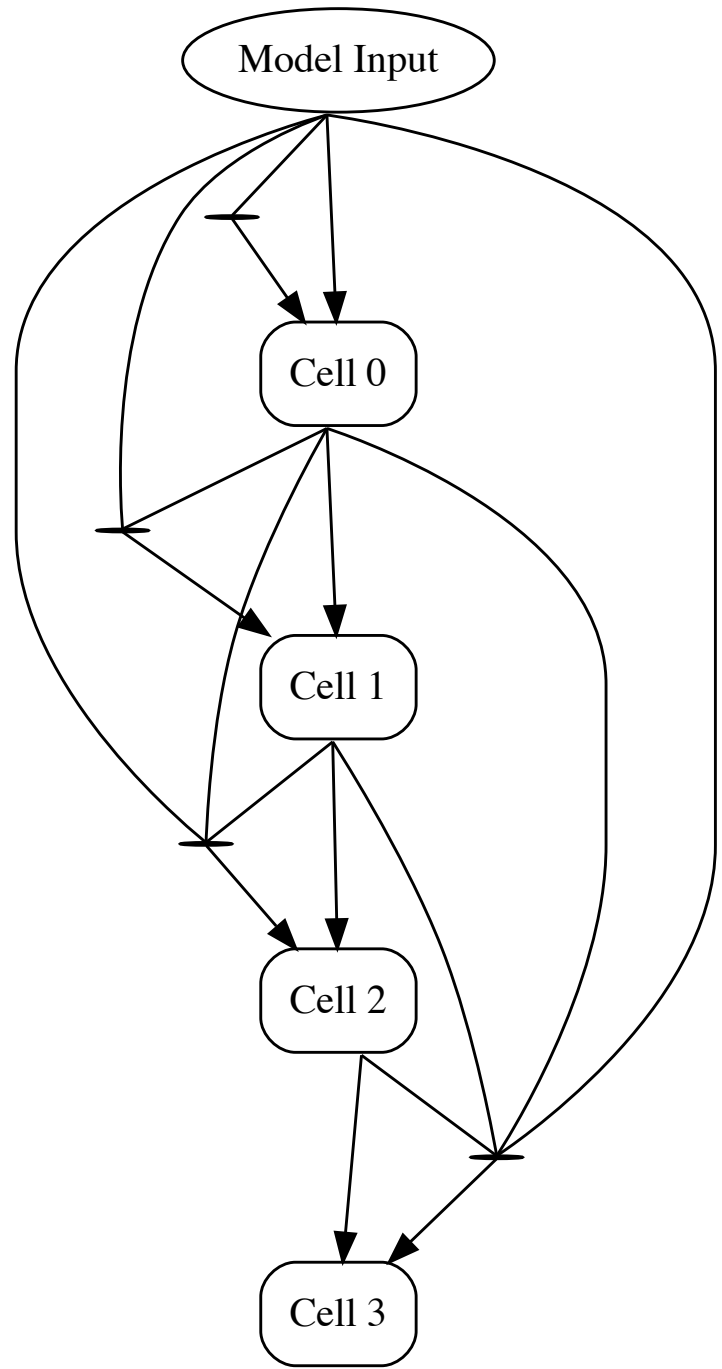

(b) The intercellular connectivity for a small four cell model. Pruners are again represented by the horizontal black boxes. Each cell receives a permanent, hard-coded input from the previous cell, as well as a prunable combination of the outputs of each of the previous cells.

Figure 2: Intra- and inter-cellular hyper-connectivity

# 3 Details of an Example Run

In Table 3 the cellular makeup, total model size, compression targets, and epochs spent training each section are listed for an example run of Bonsai-Net.

| Section | Cells | Total # Cells | $c_{target}$ | Epochs |
|---|---|---|---|---|
| 0 | [N, N] | 2 | 89.4% | $0 \to 7$ |
| 1 | [R] | 3 | 46.4% | $8 \to 51$ |
| 2 | [N, N] | 5 | 51.8% | $52 \to 59$ |
| 3 | [R] | 6 | 41.1% | $60 \to 75$ |
| 4 | [N, N] | 8 | None | $76 \to 600$ |

Table 3: The cellular makeup (where 'N' refers to a normal cell, and 'R' a reduction), total cumulative cell count, compression target, and the number of training epochs used for each section from an example Bonsai-Net run on CIFAR-10. The model compressed to 28.4% by epoch 600 solely due to classification performance.

# 4 Operation Counts

To visualize the architectural decisions made by Bonsai-Net models, here we have plotted the number and type of operations that were in each of the three Bonsai-Net models at the end of 600 epochs, grouped by cell type and by cell depth within the model.

## 4.1 By Cell Type

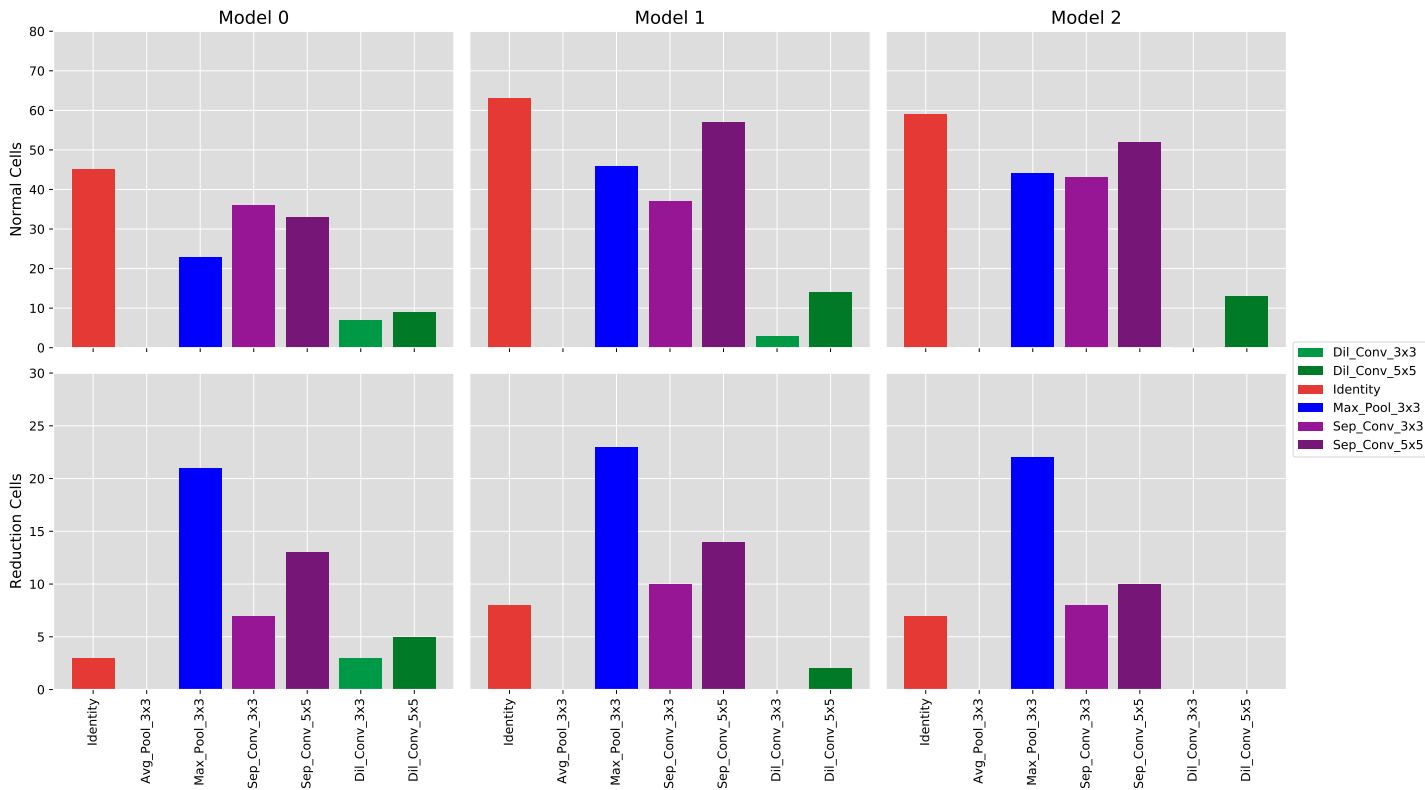

Figure 3: Here we see the total operation counts of each normal and reduction cell across three separate Bonsai-Net runs. Notice the similar operation frequencies between cells of the same type across different runs.

## 4.2 By Cell Depth

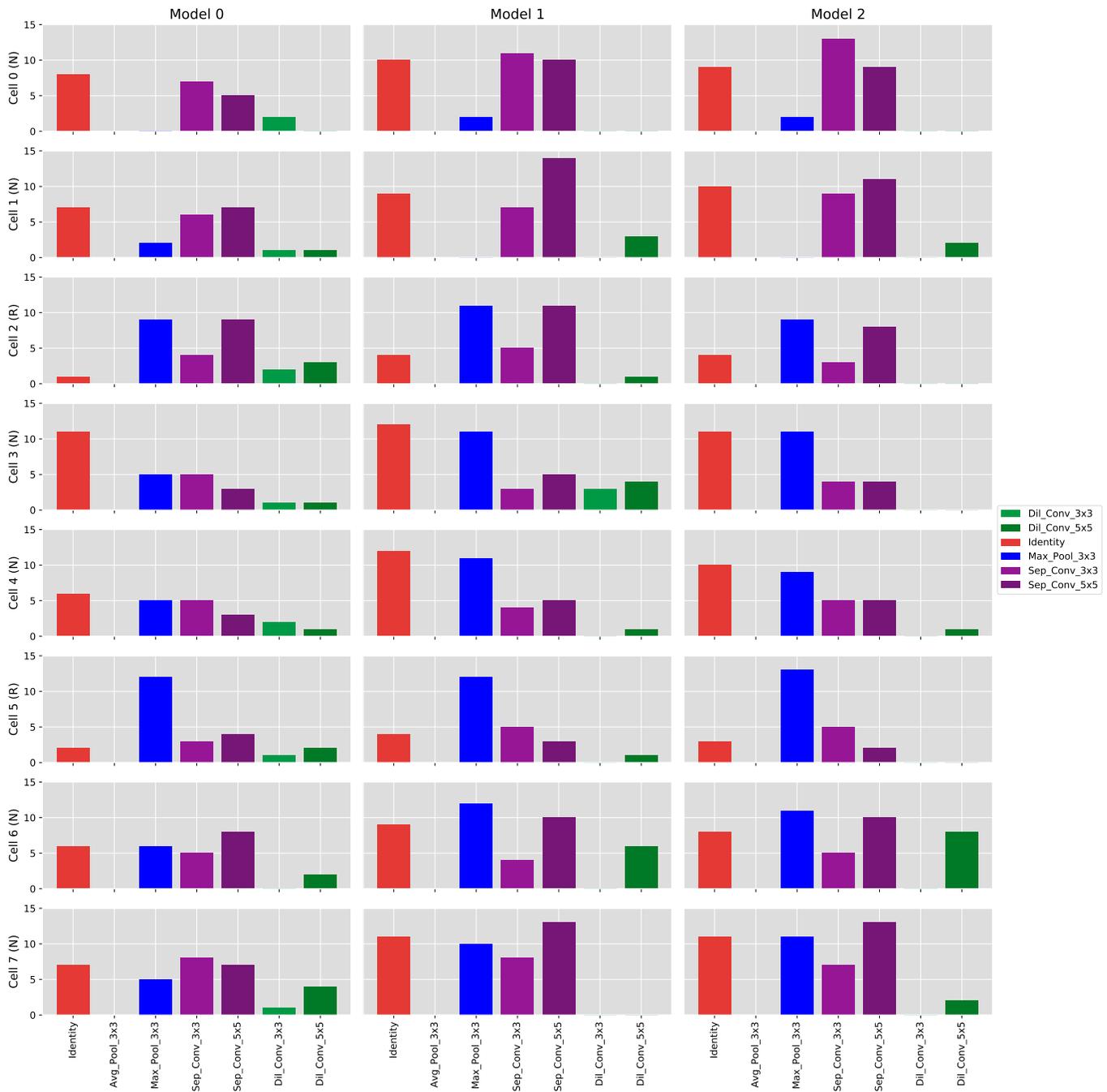

Figure 4: Here we see the total operation counts from each cell in three separate Bonsai-Net runs. Notice the frequency similarities between cells at the same depth across the three runs, as well as the drastically different schema for reduction cells.

4

# 5 Learned Intracellular Connectivity

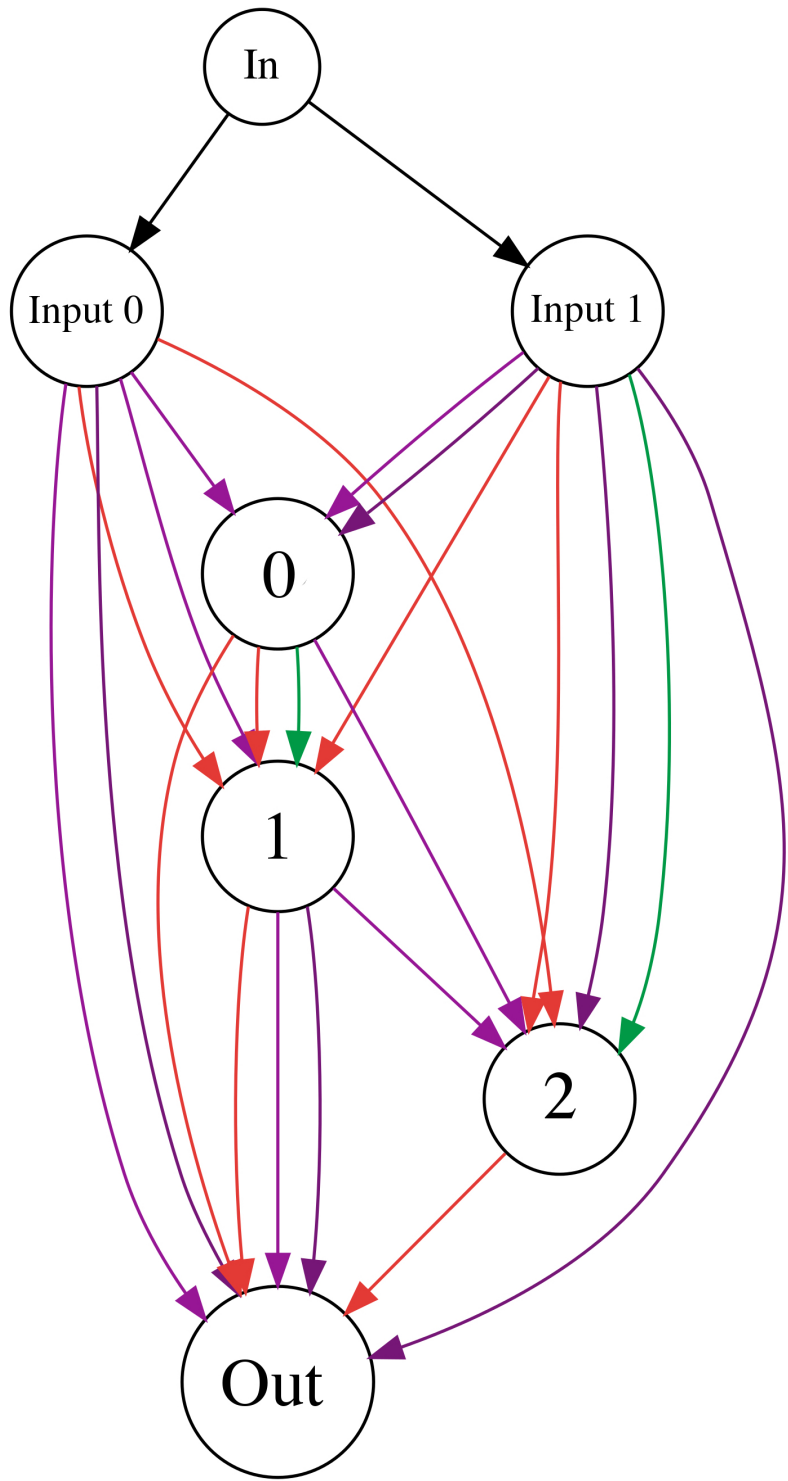

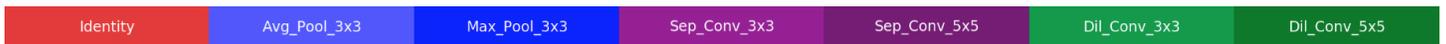

Figure 5: Above is the learned connectivity of the zeroth cell from Model 1 above, after 600 epochs of training. The 'In' node refers to the input of the model, which passes unmodified to the two input nodes of the cell, Input 0 and Input 1. Notice the ResNet-esque patterns appearing at different scales. Across two-nodes, such patterns appear as Input 1 → 1, Input 2 → 2, or 1 → Out, where identity and convolution operations run parallel with each other into a summation node. They also appear across three-node patterns, such as Input 1 → 0 → 2 or 0 → 1 → Out.



\end{document}